\documentclass[a4paper, conference]{IEEEtran}      

\IEEEoverridecommandlockouts                              

\usepackage{graphics} 
\usepackage{graphicx}
\usepackage{dblfloatfix} 
\usepackage{amsmath} 
\usepackage{amssymb}  

\usepackage[ruled,vlined]{algorithm2e}

\usepackage[utf8]{inputenc} 
\usepackage[T1]{fontenc}    
\usepackage{hyperref}       
\usepackage{url}          
\usepackage{booktabs}      
\usepackage{amsfonts}       
\usepackage{nicefrac}       
\usepackage{microtype}     
\usepackage{caption}
\usepackage{subcaption}
\usepackage{catchfile}
\usepackage{totcount}
\usepackage{multirow}
\usepackage{array}
\newcolumntype{L}{>{\centering\arraybackslash}m{1.15cm}}
\newcolumntype{P}{>{\centering\arraybackslash}m{0.75cm}}

\begin{document}

\title{\large {\bf
A Framework for Real-time Traffic Trajectory Tracking, Speed Estimation, and Driver Behavior Calibration at Urban Intersections Using Virtual Traffic Lanes} \\
{\small *accepted for publication at the 24th IEEE Intelligent Transportation Systems Conference (ITSC 2021)
\thanks{\copyright 2021 IEEE Personal use of this material is permitted. Permission from IEEE must be obtained for all other uses, in any current or future media, including reprinting/republishing this material for advertising or promotional purposes, creating new collective works, for resale or redistribution to servers or lists, or reuse of any copyrighted component of this work in other works.}}
}

\author{\IEEEauthorblockN{\bf Awad Abdelhalim, Montasir Abbas}
\IEEEauthorblockA{\textit{Department of Civil and Environmental Engineering} \\
\textit{Virginia Polytechnic Institute and State University} \\
Blacksburg, Virginia, USA \\
atarig@vt.edu, abbas@vt.edu}

\and

\IEEEauthorblockN{\bf Bhavi Bharat Kotha, Alfred Wicks}
\IEEEauthorblockA{\textit{Department of Mechanical Engineering} \\
\textit{Virginia Polytechnic Institute and State University} \\
Blacksburg, Virginia, USA \\
bharat91@vt.edu, awicks@vt.edu}
}

\maketitle
\begin{abstract}
In a previous study, we presented VT-Lane, a three-step framework for real-time vehicle detection, tracking, and turn movement classification at urban intersections. In this study, we present a case study incorporating the highly accurate trajectories and movement classification obtained via VT-Lane for the purpose of speed estimation and driver behavior calibration for traffic at urban intersections. First, we use a highly instrumented vehicle to verify the estimated speeds obtained from video inference. The results of the speed validation show that our method can estimate the average travel speed of detected vehicles in real-time with an error of 0.19 m/sec, which is equivalent to 2\% of the average observed travel speeds in the intersection of study. Instantaneous speeds (at the resolution of 30 Hz) were found to be estimated with an average error of 0.21 m/sec and 0.86 m/sec respectively for free flowing and congested traffic conditions. We then use the estimated speeds to calibrate the parameters of a driver behavior model for the vehicles in the area of study. The results show that the calibrated model replicates the driving behavior with an average error of 0.45 m/sec, indicating the high potential for using this framework for automated, large-scale calibration of car-following models from roadside traffic video data, which can lead to substantial improvements in traffic modeling via microscopic simulation.
\end{abstract}


\section{INTRODUCTION}

With over 3,000 daily fatalities, traffic crashes remain one of the major causes of death around the globe. In the United States, 40,000 lives are lost annually due to traffic crash-related fatalities, 21.5\% of which are due to urban intersection-related crashes. Intersection-related crashes comprise about 40\% of all the reported annual crashes in the USA \cite{fhwa}.

Intelligent Transportation Systems (ITS) aim to address the safety issues in transportation by revolutionizing the transport systems from being passive to smart, proactive systems. Given the increasing use of traffic cameras for monitoring and enforcement in recent years, offline traffic video data and online traffic video streams are becoming more available to researchers and practitioners. The process of accurate and efficient traffic trajectory tracking remains a challenging task. The quality of data available from video streams of on-site cameras, the full and partial occlusion of vehicles in congested traffic, heavy computational demand, and driver privacy concerns all contribute to additional complications that demand further improvements for the state-of-the-art practices and methodologies.  


\subsection{Object Detection and Tracking for Traffic Safety}

Recent years have witnessed gigantic advancements in the fields of object detection and tracking. The development and continuous improvement of Convolutional Neural Networks (CNNs) played a major role in those advancements. CNNs form the basis of the vast majority of the state-of-the-art detection and tracking methodologies. Achieving real-time performance was made possible with the introduction of Fast R-CNN \cite{frcnn} and shortly after, You Only Look Once (YOLO) and Single-Shot Detector (SSD) \cite{yolo}, \cite{ssd}. Rapid improvements in those models have followed in recent years \cite{maskcnn}, \cite{bochkovskiy2020yolov4}. In the transportation field, researchers have tried tackling the problem of vehicle detection and tracking in traffic using a myriad of methodologies including the use of adaptive background memory \cite{adaptive} and scene analysis through belief networks \cite{belief}. Tracking vehicles in congested urban conditions has been a challenge for those early methods due to occlusion. Later models utilizing sub-feature tracking were able to better handle this problem \cite{beymer1997real}, \cite{kamijo2000traffic}. The advancements in GPU-accelerated computing and deep learning methodologies enabled researchers in the field of traffic engineering to develop models that detect, track, and extract vehicle trajectories in real-time \cite{shirazi2019trajectory}, \cite{wu2017traffic}. A growing number benchmark datasets is being made available to support the research efforts in the field \cite{tang2019cityflow}, \cite{wen2020ua}, \cite{naphade20204th}.

\subsection{The Use of Object Detection and Tracking in Vehicle Speed Estimation}
Researchers have utilized different techniques using on-board equipment, optical flow, image scaling, and deep learning methods, among others. Sreedivi and Gupta \cite{indu2011optflow} used optical flow and feature point tracking in an image to estimate the velocity of the vehicle using MATLAB and Simulink. A method utilizing optical flow was also proposed and assessed by Dogan et. al. \cite{dougan2010real}. Osamura et. al. \cite{osamura2013onboard} proposed a method for feature point tracking from video data utilizing an on-board drive recorder. Costa et. al. \cite{costa2020imgfactor} proposed a method utilizing image scale factoring to calculate the velocities of the vehicles travelling in the longitudinal direction compared to the axis of the camera. While the proposed approach results in accurate speed estimates, it is difficult to automate since it requires knowledge of the actual width of the vehicles being tracked. Garg and Goel \cite{garg2013real} proposed a method utilizing license plate recognition. Their method resulted in speed estimates within $\pm$8 km/hr of the actual vehicle speeds. Luvizon et. al. \cite{luvizon2016video} proposed a method based on licence plate tracking and were able to estimate speeds with an average error of -0.5 km/hr. 

The recent improvements in vehicle and tracking resulted in major strides in vehicle speed estimation. Dong et. al. \cite{dong2019vehicle} utilized a method based on 3-dimensional convolutional networks to estimate the average vehicle speed based on information from video footage. The proposed method utilized non-local blocks, optical flow, and a multi-scale convolutional network. The results show that the proposed method results in reliable speed estimation with a mean absolute error of 2.71 km/h. Huang \cite{huang2018traffic} used perspective transformation to estimate vehicle speeds. While using assumptions for the pixels to field measurement conversion, the study by Huang shows that perspective transformation can result in accurate speed estimation with reduced computational complications. 

Accurate and efficient speed estimation remains a challenging task, especially from low-altitude roadside cameras where methods utilizing license plate tracking and deep neural nets outperform those based on vehicle tracking due to detection instability and frequent identity switches. Given the proven ability of our VT-Lane vehicle tracking framework to provide accurate vehicle trajectories in real-time while resolving vehicle identity switches due to occlusion \cite{abdelhalim2020vt, abdelhalim2020towards}, the authors believe there is a high potential for speed estimation with reliable accuracy from the produced trajectories without the need for additional complex deep neural nets or the privacy concerns associated with license plate tracking.

\subsection{Objective}
In this study, we assess an end-to-end application of our VT-Lane framework to obtain real-time vehicle trajectories and movement classification, followed by image processing and reference object scaling to obtain vehicle speed estimates. We verify the accuracy of the speed estimation using high-granularity data obtained from an instrumented vehicle that is tracked as it is driven through the study site. We conclude with a value proposition of utilizing the high accuracy speed estimates for the calibration of car-following models.

\section{METHODOLOGY}

\subsection{Vehicle Tracking and Movement Classification Framework}
For the base object detection and tracking task in VT-Lane, we implemented a combination of YOLO v4 \cite{bochkovskiy2020yolov4} and Deep-SORT \cite{wojke2017simple}. Our complete framework is detailed and assessed in \cite{abdelhalim2020vt}. The proposed framework has proven efficient in trajectory tracking, turn movement classification, and trajectory reconstruction after effectively handling identity switches. The efficacy of our work was further verified through the 2020 AI City Challenge \cite{abdelhalim2020towards}, \cite{naphade20204th}. The YOLO v4 object detector was pre-trained on the Microsoft Common Objects in Context (COCO) dataset. For this study, we improve the Deep-SORT tracker by retraining on the UA-Detrac dataset \cite{wen2020ua} which resulted in significant performance improvements in congested traffic conditions. We also incorporated Hou et. al's work \cite{hou2019vehicle} for low confidence track filtering. This pipeline runs at $\sim$35 fps on NVIDIA Tesla V100 GPU. Given that the majority of traffic control devices run at a 10 Hz rate, the real-time capabilities of this pipeline ensure seamless integration with the existing infrastructure.

\subsection{Image Processing and Speed Estimation}

\begin{figure}[!h]
\centering
\includegraphics[width = 0.48\textwidth, height = 1.4in]{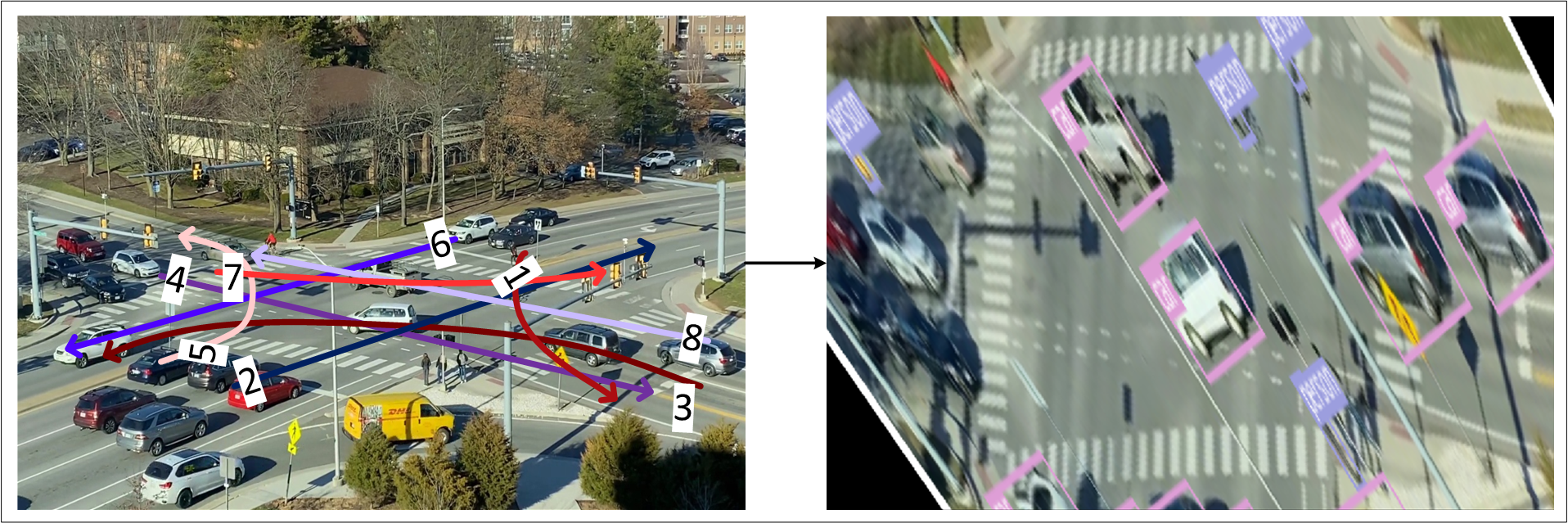}
\caption{\label{fig:detection} Original video view and perspective transformation.}
\end{figure}

\begin{figure}[!h]
\centering
\includegraphics[scale=0.25]{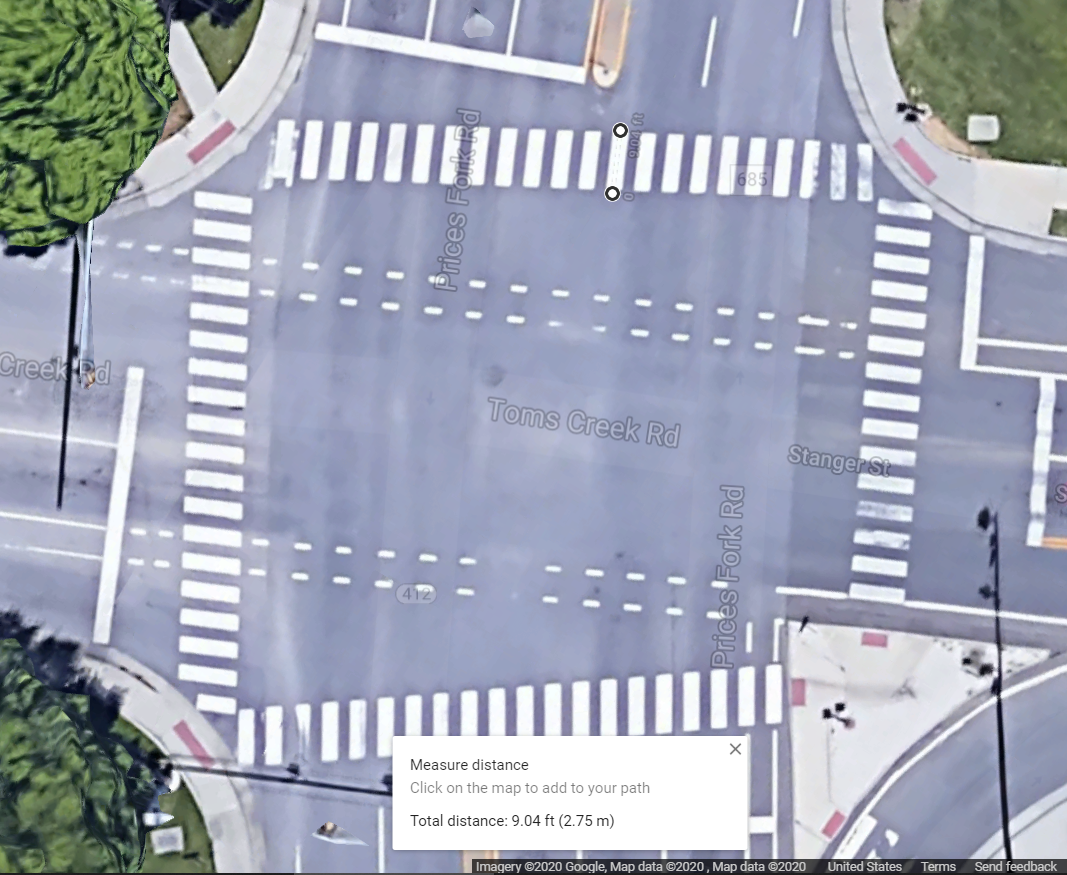}
\caption{\label{fig:google} Google Maps image of the study intersection.}
\end{figure}

Detection and tracking of vehicles takes place in the original camera space. We then utilize perspective transformation to flatten the traffic intersection scene of the area of study into a bird's-eye 2D space as shown in Figure \ref{fig:detection}, which also illustrates the NEMA movements for the site of study. The transformed perspective results in a similar view to that of the satellite image of the intersection as shown in Figure \ref{fig:google}.


We take advantage of this linearized transformed perspective to calculate the meters per pixel (MPP) ratio using reference objects, which allows the calculation of the actual distance traveled by vehicles. We used the pedestrian crossing markings around our region of interest inside the intersection as a reference, being of standard length on all sides of the intersection. The markings are $\sim 2.75$ meters in length. The lengths of the markings in the transformed perspective space were found to change from 130 to 90 from the bottom to the top in the Y direction, respectively, and 60 to 100 from the left to the right in the X direction, respectively.

To account for the change in MPP, when calculating distances for speed calculations, we use euclidean distance with applied weights to the changes in MPP for the X and Y coordinates as shown in Equation \ref{eq:weighted_euclidean}. A linear interpolation function is applied to calculate the average MPP respectively for X and Y coordinates based on the midpoint of $\Delta$ X and $\Delta$ Y. For instantaneous speed calculation, $\Delta$ X and $\Delta$ Y would respectively be the frame-to-frame change in the location of the centroid of the detected vehicle. Knowing the actual distance traversed and the framerate of the input video, the frame-by-frame speed is calculated following Equation \ref{eq:speed}.

\begin{equation}
    Distance_{point_1,\ point_2} = \sqrt{(\Delta X*W_x)^2 + (\Delta Y*W_y)^2}
  \label{eq:weighted_euclidean}
\end{equation}

\noindent Where: \\
$Distance$ = Weighted Euclidean Distance in $m$.\\
$X$ and $Y$ = Pixel coordinates in transformed perspective space of the two points (in this case, the centroids of bounding boxes for the detected vehicles).\\
$W_x$ and $W_y$ = Average MPP in the X and Y directions, respectively.

\begin{equation}
  \footnotesize W_c = \frac{MPP_{c\ max} - MPP_{c\ min}}{c_{max} - c_{min}} * (c_{midpoint} - c_{min}) + MPP_{c\ min}
  \label{eq:weights}
\end{equation}

\begin{equation}
  c=
  \begin{cases}
    x, & \text{for}\ X\ coordinates \\
    y, & \text{for}\ Y\ coordinates
  \end{cases}
  \label{eq:cases}
\end{equation}

\noindent Where: \\
$W_c$ = Weight of coordinates in meters/pixel.\\
$MPP_{c\ min}$ and $MPP_{c\ max}$ = The meters/pixel for the coordinates at the minimum (left for X, bottom for Y) and maximum (right for X, top for Y), respectively.\\
$c_{min}$ and $c_{max}$ = The minimum and maximum pixel location for the coordinate axis, respectively.\\
$c_{midpoint}$ = The midpoint of the distance being calculated. \\


The estimated speed profile of all vehicles is calculated by applying Equation \ref{eq:speed} to all vehicles across all detected frames as shown in Algorithm \ref{algorithm:speed}. The speed estimation is skipped for the first appearance of each vehicle as there is no prior knowledge of the location of the vehicle in previous time steps.

\begin{equation}
  \footnotesize Speed_{i,k} = \frac{Distance(centroid_{i,\ k}\ ,\ centroid_{i,\ k-1})}
  {(frame_k - frame_{k-1})/fps} \hspace{20pt} \forall_{i,\ k}
  \label{eq:speed}
\end{equation}

\noindent Where: \\
$Speed_{i,k}$ = The estimated speed of vehicle $i$ in frame $k$ in $meter/sec.$\\
$centroid_{i,k}$ and $centroid_{i,k-1}$ = The centroids of the bounding box for tracked vehicle $i$ in the current and previous frame in which it was detected, respectively.\\
$frame_k$ and $frame_{k-1}$ = The current and previous frame in which vehicle $i$ was detected, respectively.\\
$fps$ = The number of frames per second (=30 for this study).


\begin{algorithm}
\SetAlgoLined
\KwResult{Estimating the frame-by-frame speed for all identified vehicles.}
initialization\;
\For{$car_i$}{
    \For{$frame_k$}{
        \eIf{$frame_{i,k}$ \ = \ $frame_{i,1}$}{
            Continue\;
            }{$
            Speed_{i,k} \leftarrow{} \frac{Distance(centroid_{i,\ k}\ ,\ centroid_{i,\ k-1})}{(frame_k - frame_{k-1})/fps}
            $\;
            }
        }
    }
\caption{Speed Estimation}
\label{algorithm:speed}
\end{algorithm}

We utilize a highly instrumented vehicle driving through the study site to validate the speeds. The instantaneous speed of the instrumented vehicle is estimated as it passes through the camera frame, which is then compared to the instantaneous speed recorded by the vehicle's on-board devices.

\subsection{GHR Model Calibration}
Following speed estimation and validation, and identification of car-following episodes based on the vehicles' NEMA movement classification, we assess the value of utilizing the instantaneous speed estimates for calibrating car-following models. For this purpose, we utilize the Gazis-Herman-Rothery (GHR) car-following model. Previous studies show that the GHR model generalizes well \cite{higgs2014segmentation}, \cite{abdelhalim2020vehicle}. Algorithm \ref{algorithm:speed} shows the steps followed for calibrating the GHR model for each NEMA phase for the intersection of study (Figure \ref{fig:detection}). 
\begin{algorithm}
\SetAlgoLined
\KwResult{Estimating the GHR speed and error for all instances of all car-following episodes within each NEMA movement in the intersection.}
 initialization\;
  \For{$NEMA\_Phase_i$}{
    \For{$episode_j$}{
        \For{$instance_k$}{
            \eIf{$k \ \leq \ Floor(\frac{T}{1/fps})$}{
                Continue\;
                }{
                $\hat{a}_{j,k} \leftarrow{} cv_n^m(t)\frac{\Delta v(t-T)}{\Delta x^l(t-T)}$\;
                $\hat{v}_{j,k} \leftarrow{} \hat{v}_{j,k-1} +  \frac{\hat{a}_{j,k} + \hat{a}_{j,k-1}}{2*1/fps}$\;
            $error_{j,k} \leftarrow{} |v_{j,k} - \hat{v}_{j,k}|$\;
        }
        }  
        }
        $MAE_i \leftarrow{} \frac{1}{n_{phase\_instances}} \sum_{j=1}^{n_{episodes}}\sum_{k=1}^{n_{instances}} error_{j,k}$
        }
\caption{GHR Model Speed Estimation}
\label{algorithm:speed}
\end{algorithm}

The GHR model estimates the instantaneous acceleration of the following vehicle in a car-following episode using the formula below:

\begin{equation}
    a_n(t) = cv_n^m(t)\frac{\Delta v(t-T)}{\Delta x^l(t-T)}
\end{equation}

\noindent Where: $c,\ m,\ l, \& \ T$ are GHR model parameters, $a_n(t)$ and $v_n(t)$ are the instantaneous acceleration and speed of vehicle $n$ at time $t$, respectively, and $\Delta v\ \& \ \Delta x$ are the differences in speed and distance traveled between the leading and following vehicles in a car-following episode.


For the following vehicle in each episode$_j$ of phase$_i$, the GHR model acceleration $\hat{a}$ during instance$_k$ of the episode is calculated based on the distance and speed difference between the leading and following vehicles in the episode at time $(t-T)$ where $T$ is the following driver's perception-reaction time. An estimated speed $\hat{v}$ is calculated based on the GHR model acceleration. The calculated instantaneous speed for the following vehicle $\hat{v}$ is compared to the speed $v$ estimated by our algorithm (considered to be ground truth speed in this case). An optimization problem is formulated and solved to find the GHR model parameters that minimize the error in terms of the Mean Absolute Error for each NEMA movement (MAE$_i$) subject to: $\scriptstyle 0.2 \leq T \leq 1.8, \ 0 \leq m \leq 3, \ 0 \leq l \leq 2, \ 0 \leq c \leq 3$.




\section{INSTRUMENTED VEHICLE CONFIGURATION}
\begin{figure}[!h]
\centering
\includegraphics[width= 0.48\textwidth]{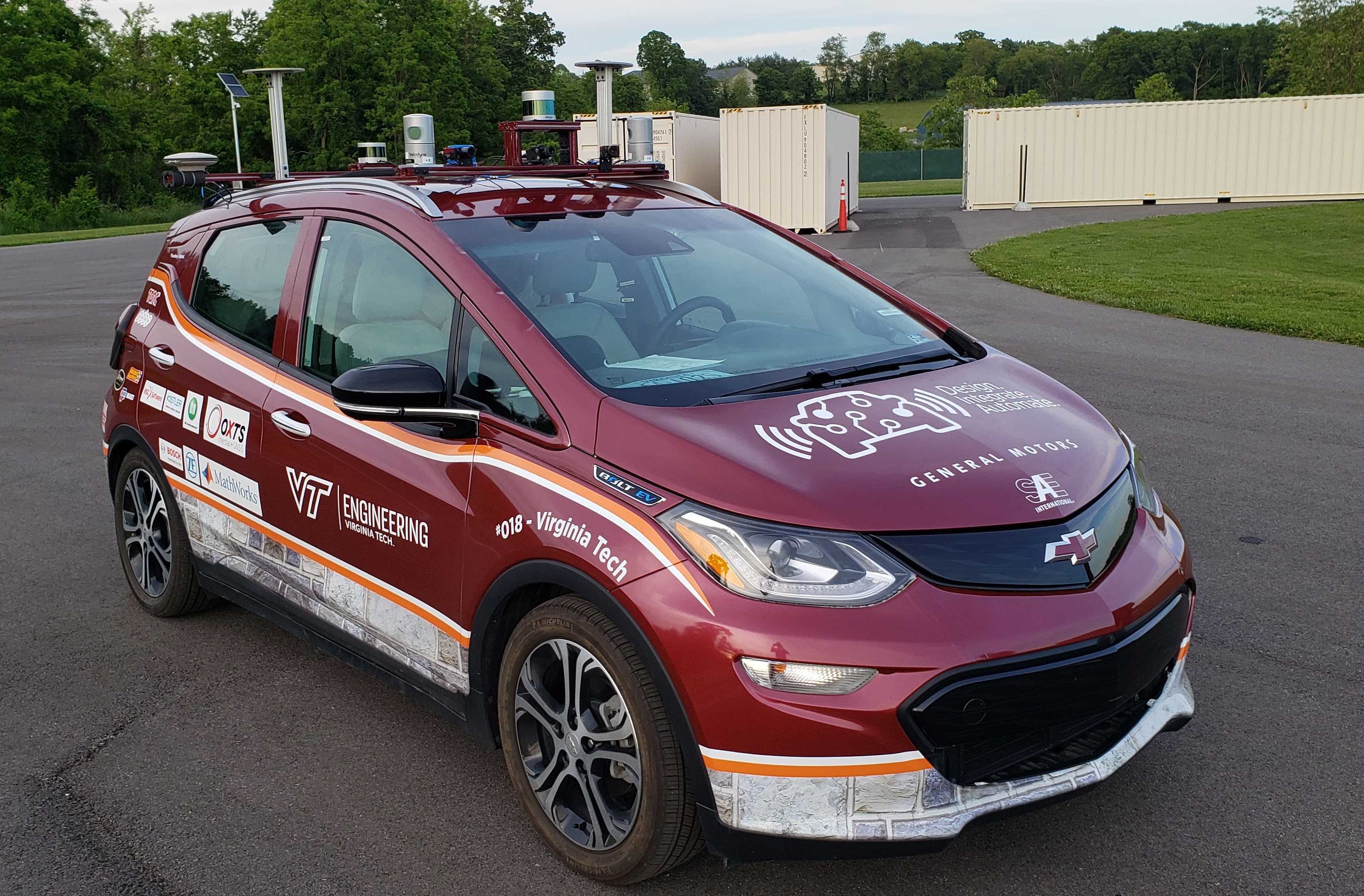}
\caption{\label{fig:autodrive_car} The instrumented vehicle used in this study.}
\end{figure}

 The instrumented vehicle that has been used to test and validate our framework's speed estimation is shown in Figure \ref{fig:autodrive_car}. We use a 2017 Chevrolet Bolt equipped with a sensor suite to provide real-time localization of the car. The localization sensor suite of the car consists of a TORC Robotics Pinpoint Module \cite{torc} and a computing platform (AS Rock Beebox). The TORC pinpoint module is internally equipped with an Inertial Measurement Unit (IMU), its own small internal computing platform and has two external Novotel GPS sensors attached to it. The data from the IMU and the GPS sensors is fed into the computing platform of Pinpoint and processed using an extended Kalman filter to provide an update on the pose of the car at a rate of 100 Hz.
 
\begin{figure}[!h]
\centering
\includegraphics[width= 0.48\textwidth]{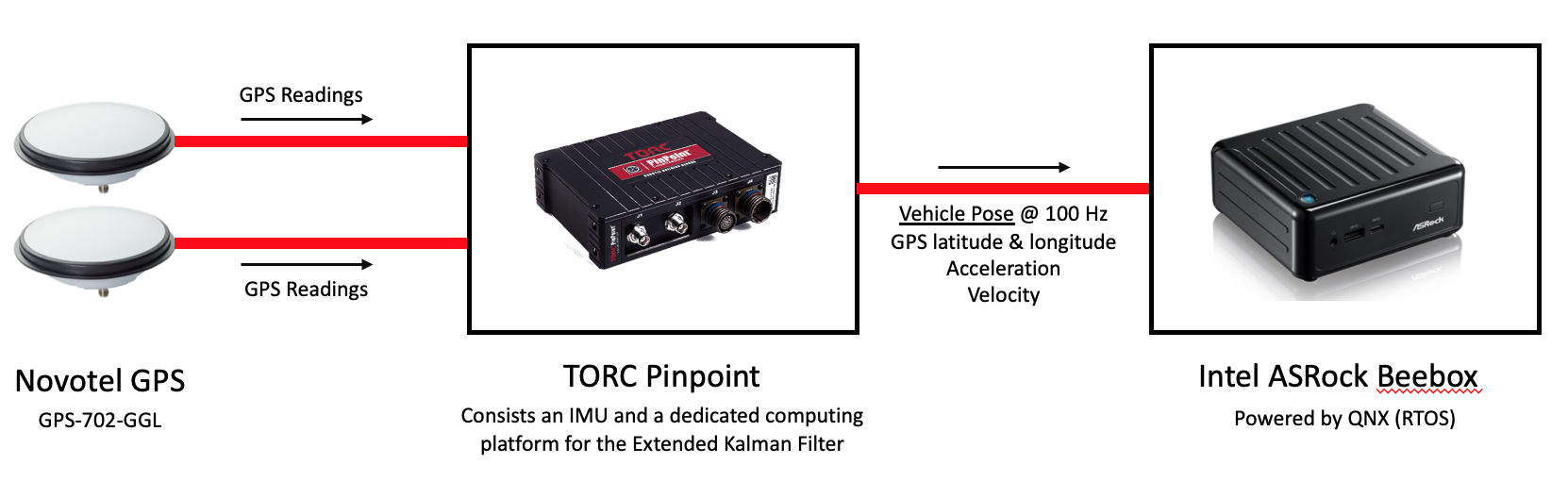}
\caption{\label{fig:car_sensors} Block diagram of the sensor suite in the car.}
\end{figure}

The pose information of the car consists of the location of the car in global coordinates (i.e. latitude and longitude), the acceleration and the velocity of the car. This output from pinpoint is read in at a 100 Hz by our computing platform (Beebox) in the car which is powered by QNX - a real time operating system. QNX is time-synced with the TORC Pinpoint to avoid any time discrepancies. Figure \ref{fig:car_sensors} illustrates the flow of the information between the sensors in the car.

\begin{figure*}[!t]
\centering
\includegraphics[width = \textwidth, height= 3.2in]{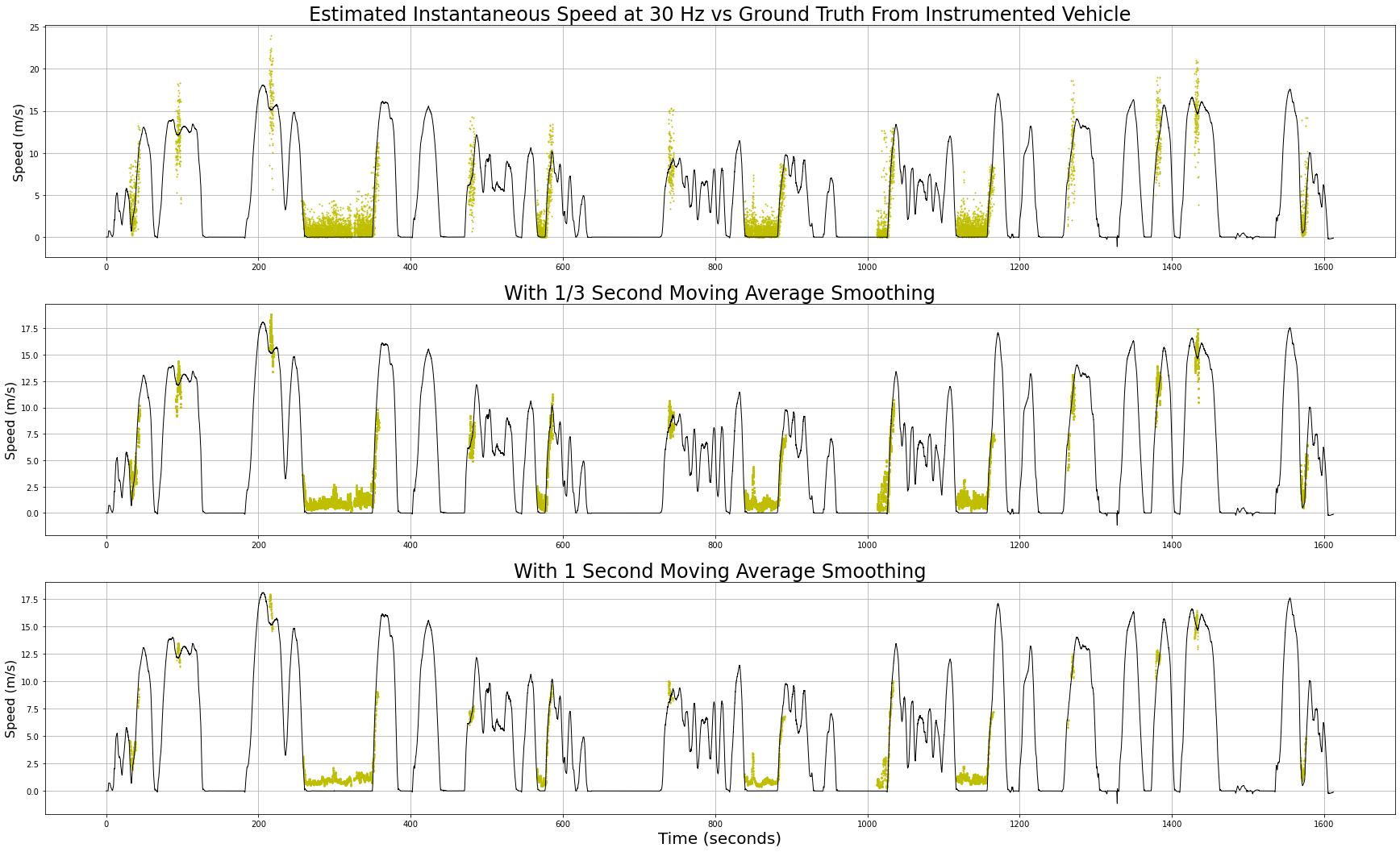}
\caption{\label{fig:speed_profile} Instrumented vehicle's speed profile (at 100 Hz) and estimated speeds with varying moving average window sizes.}
\end{figure*}

\vspace{-15pt}
\section{RESULTS AND ANALYSIS}

\subsection{Speed Estimation and Validation}
A 1-hour video was recorded for the intersection shown in Figure \ref{fig:detection} during the PM peak hour. The instrumented vehicle was driven to make different movements through the intersection during that time. All vehicles maneuvering the intersection were tracked, and classified based on their NEMA phase movement (also illustrated in Figure \ref{fig:detection}). Speed was estimated for each vehicle on frame-by-frame basis. We used the speed data collected from the instrumented car to compare to the estimated speeds when the car appeared and was tracked in the camera frame within the area of study. The car was also driven during off-peak time to make uninterrupted movements through the intersection. This was made so as to compare the performance of our speed estimation pipeline when there is no traffic occlusion and stop-start movement at the intersection. Table \ref{tab:average} shows the average ground truth speeds during 15 different appearances for the instrumented vehicle at congested and uncongested traffic conditions, and the estimated average speed for those appearances.

\begin{table}[!h]
  \centering
  \renewcommand{\arraystretch}{1.0}
  \caption{Average Ground Truth vs Estimated Speeds}
    \begin{tabular}{lLLLLL}
    \toprule
          & \textbf{Traffic State} & \textbf{NEMA Phase} & \textbf{Car Speed (m/sec)} & \textbf{Estimated Speed (m/sec)} & \textbf{Error (m/sec)} \\
    \hline
    1     & Light & 2 & 8.81  & 9.37  & +0.56 \\
    \hline
    2     & Light & 6 & 10.87 & 10.39 & -0.48 \\
    \hline
    3     & Light & 2 & 10.61 & 11.10 & +0.49 \\
    \hline
    4     & Light & 5  & 7.04  & 7.22  & +0.18 \\
    \hline
    5     & Light & 7  & 5.83  & 6.21  & +0.38 \\
    \hline
    6     & Light & 1  & 6.60  & 7.16  & +0.56 \\
    \hline
    7     & Light & 7  & 5.84  & 5.87  & +0.03 \\
    \hline
    8     & Congested & 3  & 9.74  & 8.35  & -1.39 \\
    \hline
    9     & Congested & 6 & 12.17 & 12.71 & +0.54 \\
    \hline
    10    & Congested & 2 & 15.20 & 16.60 & +1.40 \\
    \hline
    11    & Congested & 5  & 6.89  & 6.92  & +0.03 \\
    \hline
    12    & Congested & 7  & 8.20  & 9.00  & +0.80 \\
    \hline
    13    & Congested & 2 & 9.16  & 8.42  & -0.74 \\
    \hline
    14    & Congested & 6 & 10.81 & 11.26 & +0.45 \\
    \hline
    15    & Congested & 7  & 15.09 & 15.12 & +0.03 \\
    \bottomrule
    \multicolumn{3}{c}{\textbf{Average}} & 9.52 & 9.71 & +0.19 \\
    \bottomrule
    \end{tabular}%
  \label{tab:average}%
\end{table}%

The results indicate a very high accuracy in estimating the average speed of the vehicle, with an average error of 0.19 m/sec. The results in Table \ref{tab:average} also show that the speed estimation seems to be significantly better in light traffic.


The ground truth speed profile of the instrumented car for a part of the PM peak hour video is shown in Figure \ref{fig:speed_profile}. The yellow scatter points illustrate the estimated instantaneous speed as the instrumented car passes through the camera frame. It can be clearly noticed that there is plenty of noise in case of the raw, unfiltered estimates. We use a moving average smoothing technique to provide more reliable estimates. Figure \ref{fig:speed_profile} shows the estimated speeds using 1/3-second and 1-second moving average smoothing. While smoothing over longer periods of time would provide improved results, it can be seen that 1-second smoothing is sufficient to provide very reliable speed estimates for the instrumented car. Table \ref{tab:stats} shows the descriptive statistics of instantaneous speed estimation during the PM peak hour and off-peak times, where N is the number of instances (frames) for which the instantaneous speed of the instrumented car was estimated. 

\begin{table}[!h]
  \centering
  \renewcommand{\arraystretch}{1.3}
  \caption{Descriptive Statistics of Estimated Speeds}
    \begin{tabular}{lPPPPPP}
    \toprule
    \textbf{Traffic} & \textbf{N} & \textbf{Moving Avg} & \textbf{$\overline{error}$ (m/sec)} & \textbf{$\sigma$ (m/sec)} & \textbf{MAE (m/sec)} & \textbf{RMSE (m/sec)} \\
    \midrule
    \multirow{3}[4]{*}{Light} & \multirow{3}[4]{*}{1,594} & None  & 0.61  & 3.10  & 2.10  & 3.16 \\
\cmidrule{3-7}          &       & $\frac{1}{3}$ sec & 0.21  & 0.80  & 0.59  & 0.78 \\
\cmidrule{3-7}          &       & 1 sec & 0.21  & 0.58  & 0.28  & 0.45 \\
    \midrule
    \multirow{3}[4]{*}{Congested} & \multirow{3}[4]{*}{8,919} & None  & 0.95  & 2.07  & 0.36  & 1.14 \\
    \cmidrule{3-7}          &       & $\frac{1}{3}$ sec & 0.78  & 0.76  & 0.16  & 0.45 \\
\cmidrule{3-7}          &       & 1 sec & 0.86  & 0.46  & 0.16  & 0.42 \\
    \bottomrule
    \end{tabular}%
  \label{tab:stats}%
\end{table}%

\begin{figure}[!h]
\centering
\begin{subfigure}{.48\textwidth}
  \includegraphics[width = \textwidth]{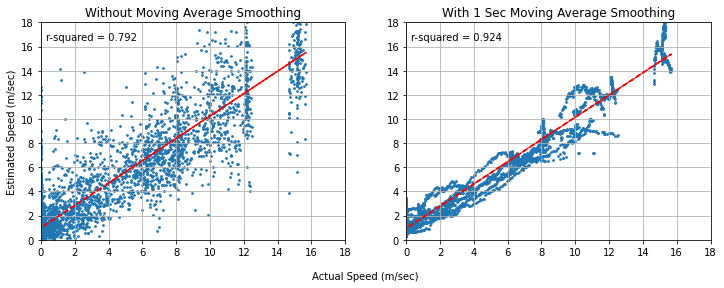}
  \caption{\label{fig:sub1} Congested Traffic Condition.}
  \vspace{5pt}
\end{subfigure}
\begin{subfigure}{.48\textwidth}
  \includegraphics[width = \textwidth]{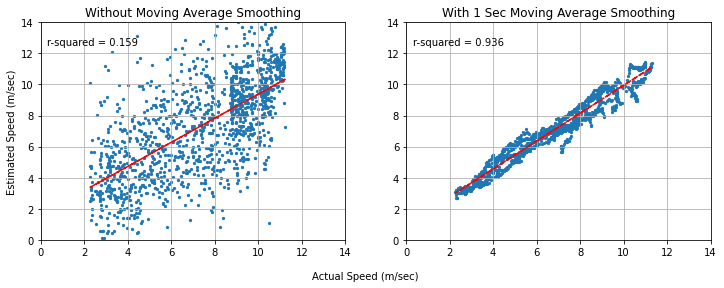}
  \caption{ \label{fig:sub2} Light Traffic Condition.}
\end{subfigure}
\caption{\label{fig:actual_pred} Actual vs Estimated Instantaneous Speed.}
\end{figure}

Figure \ref{fig:actual_pred} shows the actual vs estimated plot for those instantaneous speeds. This illustrates the high accuracy of our method in estimating the speeds when using 1-sec smoothing, as supported by the aforementioned statistics. The speed prediction during light, free-flowing is significantly more accurate than the congested state. There is one instance where a sudden, "v" shaped decrease followed by an increase in the speed of the instrumented car (second-to-last appearance in Figure \ref{fig:speed_profile}) which was not captured by our algorithm, resulting in an overestimation of the instantaneous speeds during that time which can be seen in Figure \ref{fig:actual_pred} (a) for the speeds between 15-16 m/sec. This is attributed to the significantly higher refresh rate of the speed reading from the instrumented car (100 Hz) compared to the input video that provides data to our framework at 30 Hz. Speeds were also found to be over-estimated when vehicles stop at the intersection as a result of the instability of the bounding detection box induced by occlusion as the moving traffic passes in front of idle vehicles.

\subsection{Driver Behavior Calibration}
We conclude with a value proposition of using the speed estimates obtained from video inference to calibrate driver behavior models. Following speed estimation, car-following episodes are identified based on the NEMA phase classification of the vehicle's movement across the intersection. Pairs of vehicles that are simultaneously tracked moving in the same lane of a specific NEMA movement are considered to be in car-following episodes. Table \ref{tab:driver_behavior} shows the GHR model parameters obtained following the optimization process to reduce the MAE of GHR-estimated speeds vs the estimated speeds from video inference for each NEMA phase.

\begin{table}[htbp]
  \centering
  \renewcommand{\arraystretch}{1.10}
  \caption{Calibrated GHR Model Parameters For The Intersection of Study}
    \begin{tabular}{LPPccccP}
    \toprule
    Min Instances & $\overline{D}$ (sec) & NEMA Phase & T     & m     & l     & c     & MAE (m/sec) \\
    \hline
    \multirow{8}[16]{*}{30} & \multirow{8}[16]{*}{2.63} & 2     & 1.74  & 0.10  & 2.0  & 1.39  & 0.11 \\
\cmidrule{3-8}          & & 5     & 1.06  & 0.10  & 1.03  & 1.39  & 0.25 \\
\cmidrule{3-8}          & & 4     & 1.42  & 2.26  & 0.98  & 0.18  & 0.41 \\
\cmidrule{3-8}          & & 7     & 1.17  & 0.10  & 0.10  & 1.19  & 0.31 \\
\cmidrule{3-8}          & & 6     & 0.56  & 0.30  & 0.41  & 1.77  & 0.71 \\
\cmidrule{3-8}          & & 1     & 0.62  & 0.75  & 2.00  & 3.00  & 0.47 \\
\cmidrule{3-8}          & & 8     & 1.08  & 0.10  & 2.00  & 0.10  & 0.34 \\
\cmidrule{3-8}          & & 3     & -     & -     & -     & -     & - \\
    \hline
    \multirow{8}[16]{*}{45} & \multirow{8}[16]{*}{3.10} & 2     & 1.42  & 0.10  & 2.00  & 2.07  & 0.32 \\
\cmidrule{3-8}          & & 5     & 0.77  & 0.10  & 1.27  & 1.98  & 0.38 \\
\cmidrule{3-8}          & & 4     & 1.29  & 2.30  & 1.15  & 0.23  & 0.52 \\
\cmidrule{3-8}          & & 7     & 0.66  & 0.41  & 0.10  & 0.42  & 0.51 \\
\cmidrule{3-8}          & & 6     & 0.93  & 0.10  & 0.10  & 1.75  & 0.73 \\
\cmidrule{3-8}          & & 1     & -     & -     & -     & -     & - \\
\cmidrule{3-8}          & & 8     & 1.57  & 0.10  & 2.00  & 0.98  & 0.24 \\
\cmidrule{3-8}          & & 3     & -     & -     & -     & -     & - \\
    \hline
    \multirow{8}[16]{*}{60} & \multirow{8}[16]{*}{3.60}& 2     & 1.02  & 0.10  & 1.75  & 1.34  & 0.77 \\
\cmidrule{3-8}          & & 5     & 1.44  & 0.14  & 0.10  & 0.13  & 0.24 \\
\cmidrule{3-8}          & & 4     & 0.93  & 2.33  & 1.57  & 0.54  & 0.62 \\
\cmidrule{3-8}          & & 7     & 0.52  & 0.10  & 0.10  & 0.69  & 0.57 \\
\cmidrule{3-8}          & & 6     & 1.39  & 0.10  & 0.10  & 2.35  & 0.91 \\
\cmidrule{3-8}          & & 1     & -     & -     & -     & -     & - \\
\cmidrule{3-8}          & & 8     & 1.31  & 0.10  & 2.00  & 0.97  & 0.37 \\
\cmidrule{3-8}          & & 3     & -     & -     & -     & -     & - \\
    \bottomrule
    \end{tabular}%
  \label{tab:driver_behavior}
\end{table}

The minimum number of instances indicates the minimum number of frames where both the vehicles in a given car-following episode are tracked simultaneously and have a speed estimate from the video processing (as vehicles are not necessarily detected and tracked through consecutive frames). The average duration ($\overline{D}$) shows the average tracking duration of car-following episodes for a given minimum instances threshold as the vehicles move within the camera frame. Vehicles can be (and most likely are) engaging in car-following episodes well before and after the limited duration they are captured within the camera frame. Blanks indicate that there weren't sufficient car-following episodes for the given NEMA movement to calibrate model parameters. The minimal error obtained via calibration process indicates the high potential of utilizing video-inferred speed estimates for large-scale calibration of driver behavior models.

\section{DISCUSSION AND CONCLUSIONS}
In this study, we presented an end-to-end application of our VT-Lane framework for obtaining real-time vehicle trajectories, movement classification, and speed estimation. The estimated speeds via our framework were verified using high-granularity data obtained from an instrumented vehicle that was tracked as it was driven through the intersection of study. The results of the speed validation show that our framework can estimate speeds in real-time with an error of 0.19 m/sec for estimating the average travel speed of detected vehicles, which is equivalent to 2\% of the observed average travel speed (9.52 m/sec) through the intersection of study. Instantaneous vehicle speeds at the resolution of 30 Hz were found to be estimated with an average error of 0.21 m/sec and 0.86 m/sec for free-flowing and congested traffic conditions, respectively, with an overall $R^2$ of 93\%.

We concluded this study with a value proposition of utilizing the high accuracy speed estimates for the calibration of driver behavior models. The parameters of the Gazis-Herman-Rothery model were calibrated for each of the NEMA movements in the intersection of study, with the results showing that the calibrated model replicates the driving behavior with an average error of 0.45 m/sec. The ability of our framework to provide high accuracy, real-time speed estimates, turn movement classification, and identity switch resolution, alongside this demonstrated potential for large-scale driver behavior models calibration from video inference signifies a high practical value for a myriad of applications in traffic safety and simulation modeling.

Our future work will include further expanding the VT-Lane framework and its high accuracy, high granularity outputs for applications in automated real-time performance and safety assessment. The authors will also work to incorporate machine learning for the automated identification and scaling of reference objects from the scenery of a given input traffic video, which would further streamline the speed estimation process and add more value to the framework.

\bibliographystyle{ieeetr}
\bibliography{references}

\end{document}